# A Metamodel Structure For Regression Analysis: Application To Prediction Of Autism Spectrum Disorder Severity

*Shiyu Wang*[*] *and Nicha C. Dvornek*[*†]

[*]Department of Radiology & Biomedical Imaging, Yale School of Medicine, New Haven, CT
[†]Department of Biomedical Engineering, Yale University, New Haven, CT

## ABSTRACT

Traditional regression models do not generalize well when learning from small and noisy datasets. Here we propose a novel metamodel structure to improve the regression result. The metamodel is composed of multiple classification base models and a regression model built upon the base models. We test this structure on the prediction of autism spectrum disorder (ASD) severity as measured by the ADOS communication (ADOS_COMM) score from resting-state fMRI data, using a variety of base models. The metamodel outperforms traditional regression models as measured by the Pearson correlation coefficient between true and predicted scores and stability. In addition, we found that the metamodel is more flexible and more generalizable.

## 1. INTRODUCTION

Metamodeling or meta-learning [1] often describes the process of learning from previously learned information [2]. Mitchell used the term 'bias' to describe a learning system's intrinsic preference for one generalization over another [3]. One perspective of meta-learning is that it learns to find the optimal bias for the specific task by combining multiple base learners [4]. However, the term meta-learning has no rigorous definition and its meaning differs from group to group [4]. Among all the interpretations, stacked generalization is one widely used approach in metamodeling [4, 5]. It puts all the predictions from the base models in a second space and generalizes in this second space to make the final guess for the test set [5]. Parallel learning, one of the many variations in stacked generalization, partitions a large dataset into several subsets first, and the same learner is applied to these subsets [6]. This method has been adopted in problems with large datasets such as financial time series forecasting [7].

Since neuroimaging datasets are generally small, dividing them into even smaller sub-datasets for parallel learning might not be wise. Therefore, we focused on extracting different knowledge from the same set of data and integrating these results to form the final regression prediction. Our metamodel is composed of several base models followed by a meta-level multilayer perceptron model. The base models are binary classification models and the predicted classification scores would then be fed as input into the meta-level multilayer perceptron model to generate the regression prediction.

Here, we tested 5 common classification networks as our base models: long short-term memory network (LSTM) [8], support vector machine (SVM) [9, 10], random forest (RF) [9, 11], multilayer perceptron (MLP) [9] and logistic regression (LR) [10]. The metamodel structure is tested on the public ABIDE dataset [12] with 8-fold cross-validation and new data generalization experiments. Our results show that the metamodel outperforms the traditional regression models for every base algorithm in various aspects.

## 2. METHOD

### 2.1. Metamodel structure

The metamodel structure (Fig. 1, left) can be explained as follows:

Step 1: Split the data for learning into a training set (TR) and validation set (VS). Data augmentation could be applied after the split if necessary.

Step 2: Train n base classification models {$B_1$, $B_2$, ..., $B_n$} using TR, and optimize hyperparameters using VS.

Step 3: Feed the same TR to the trained base models to generate n classification scores for each sample. Use these scores as input for the regression meta-level model, a fully connected neural network. Train the meta-level model using the TR, and optimize hyperparameters according to VS results.

Step 4: Predict the score on new, unseen data, e.g., a test set (TS), using the trained base models and meta-level model.

### 2.2. Generalizing to new data

This metamodel structure could be generalized to a new dataset (Fig. 1, right) in two methods:

Method 1: Apply the previously trained base models to the new dataset, then feed these scores to the previously

Corresponding author email: nicha.dvornek@yale.edu

trained meta-level model to generate the prediction.

Method 2: Split the new dataset into TR-new, VS-new and TS-new, with data augmentation applied after the split if desired. Use the previously trained $\{B_i\}$ to generate classification scores for TR-new. Train a new meta-level model with the TR-new classification scores, and optimize hyperparameters using VS-new. Apply the previously trained $\{B_i\}$ on TS-new and analyze TS-new classification scores with the newly trained meta-level model to get the prediction.

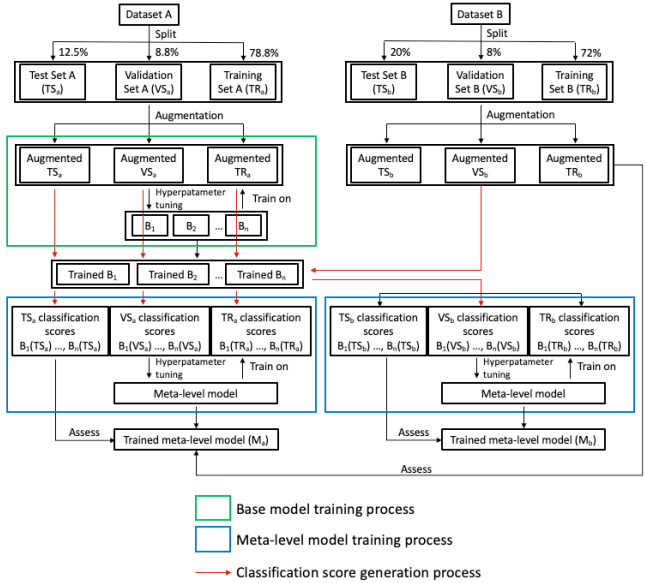

**Fig. 1**. Metamodel flowchart.

### 2.3. Advantages of the metamodel

A metamodel structure can be beneficial to the regression task for several reasons. First, each base learner is assigned with a much simpler and more targeted task by breaking down the regression problem. Instead of capturing the features needed to be able to predict the whole range of ASD severity scores all at once for traditional regression models, the base model, for example at a threshold of 0.5, would only need to learn the characteristics of samples with the score of 0 and make the binary decision. It is more flexible in the sense that each base model could get its own set of hyperparameters to better grasp the underlying features. Second, the meta-level model could serve to evaluate the performance of each base model and adjust the weights and biases on the classification scores accordingly. Third, the base model does not need to be constrained to only one type of model; different learners can be applied here as base models.

**Table 1**. Dataset A characteristics

| Institution | Number of subjects | Total number of time points | Age mean(SD) | FIQ mean(SD) |
|---|---|---|---|---|
| USM | 89 | 235 | 21(6.57) | 105(16.99) |
| NYU | 79 | 175 | 15(6.97) | 108(16.62) |
| Pitt | 26 | 195 | 19(6.80) | 111(13.53) |
| Olin | 18 | 205 | 16(3.13) | 113(17.87) |
| CMU_b | 8 | 315 | 27(6.88) | 117(8.67) |
| SBL | 8 | 195 | 33(8.53) | – |

**Table 2**. Dataset B characteristics

| Institution | Number of subjects | Total number of time points | Age mean(SD) | FIQ mean(SD) |
|---|---|---|---|---|
| CMU_a | 5 | 225 | 26(6.57) | 133(16.47) |
| KKI | 17 | 151 | 10(1.46) | 97(15.12) |
| MaxMun_d | 5 | 195 | 10(2.00) | 109(8.57) |
| SDSU | 13 | 175 | 15(1.79) | 114(16.42) |
| Stanford | 17 | 175 | 10(1.62) | 114(18.72) |
| UCLA_1 | 41 | 115 | 13(2.62) | 103(13.13) |
| UCLA_2 | 12 | 115 | 13(1.95) | 92(12.40) |

## 3. EXPERIMENTS

### 3.1. Participants

We use data from the ABIDE dataset [12]. It is an open ASD neuroimaging dataset, including resting-state fMRI (rsfMRI) and phenotypic data. We utilize the public preprocessed ABIDE data [13], choosing the Connectome Computation System (CCS) pipeline with band-pass filtering and without global signal regression and the AAL brain parcellation, which creates 116 regions of interest [13]. We specifically chose the data from two sets of institutions; the characteristics of these two datasets A and B are shown in Table 1 and Table 2.

### 3.2. Preprocessing methods

We performed an 8-fold cross validation on dataset A and split the data into 78.7% training set ($TR_a$), 8.8% validation set ($VS_a$) and 12.5% test set ($TS_a$) with a fixed random state such that the same splits are used for all models. Next, we augmented our data by extracting sequences of 90 time points with a stride of 10 time points between samples. We use the average of all sample predictions from a given subject as the predicted score for that subject.

For LSTM models, we used the rsfMRI ROI time-series combined with phenotypic data (age and FIQ) as the input features. Each phenotypic variable is z-normalized and replicated to match the dimension along the time-domain of the 116 rsfMRI features [14]. The FIQ for SBL site subjects is set to zero since their FIQ information is not available. For SVM, RF, LR and MLP, we used functional connectivity between each pair of ROIs as the input features. We computed functional connectivity as the matrix of Pearson correlation coefficients, took the upper triangle values and reshaped them into a one-dimensional vector of length 6670 as the input.

## 3.3. Methods implementation and evaluation

Our objective is to predict ASD severity from rsfMRI. ASD severity is often assessed using the Autism Diagnostic Observation (ADOS) [15], and we choose the ADOS communication score (ADOS_COMM) as our prediction target, which ranges from 0 to 8. Each base model classifies whether a sample has a score lower or higher than a given threshold. Seven thresholds 0.5, 1.5, ..., 6.5 are chosen for the base models ($B_1$, $B_2$, ..., $B_7$). Only seven cutoffs are adopted because very few patients have a score of 8, thus a base model with a cutoff of 7.5 would learn little to no information. All based models are trained on $TR_a$ with hyperparameters tuned using $VS_a$.

The LSTM base models are single layer models with 16 or 32 hidden nodes. MLP base models have two hidden layers with 1000 and 100 nodes. The loss function for both models is binary cross-entropy. The dataset could be highly imbalanced, therefore class weight is used based on the ratio of the number of majority to minority class samples in the training set. To prevent overfitting, dropout regularization, l2 regularization and Gaussian noise added to the training targets are applied to both models. SVM, RF, and LR base models are trained in MATLAB using default parameters except as noted. SVM base models use a linear kernel, tuning the hyperparameters controlling the box constraint and kernel scale. RF base models are optimized for the number of trees, and LR base models are optimized for the strength of l2 regularization.

The meta-level model here is a fully connected neural network with a 4-node hidden layer. It is trained on $TR_a$ classification scores and optimized on $VS_a$ results. We choose sigmoid as the activation function for the hidden layer and linear activation for the output.

LSTM, MLP, SVM, RF and linear regression models ($R_a$) are trained on the exact same cross-validation splits as the metamodel structure for comparison. A 16-node LSTM regression model is used (phenotypic data included), and l2 regularization, sample weight and dropout are applied. The MLP regression model has two hidden layers with 1000 and 100 nodes. For SVM, RF and linear regression models, the hyperparameters which have been optimized are the same as the ones in the base models as described above.

To evaluate the generalization of the proposed metamodel structure, here we use a new dataset B. Two generalization methods as stated above are tested using the LSTM and MLP base algorithms. After a 5-fold cross-validation split and data augmentation on dataset B, we get training set B ($TR_b$), validation set B ($VS_b$) and test set B ($TS_b$). For the first method, we apply all 8 trained metamodel structures (the trained base models $\{B_i(TS_a)\}$ followed by the trained meta-level model $M_a$) to the entire dataset B. Second, a new meta-level model ($M_b$) is trained on $TR_b$ classification scores, optimized on $VS_b$ classification scores and assessed on $TS_b$ classification scores, which are all generated by the previously trained base models $\{B_i(TS_a)\}$. For comparison, previously trained regression models ($R_a$) are applied to the entire dataset B. In addition, new traditional regression models ($R_b$) are trained and optimized on the same training and validation set as $M_b$.

## 3.4. Results and discussion

We use Pearson correlation coefficient between the true and predicted ADOS_COMM scores on the subject level to assess the model performance. The results on dataset A show that the metamodels ($M_a$) outperform the traditional regression models for all 5 methods (Table 3). Furthermore, we performed a 2-way repeated measures ANOVA to test whether there was a difference between traditional vs. metamodel pipeline results, accounting for the repeated use of the same folds for testing the different base algorithms. We found a statistically significant effect for the regression model method (traditional vs. metamodel pipeline, $F(1, 7) = 10.49$, $p = 0.01$), indicating the mean correlation for the metamodels ($M = 0.32$, $SD = 0.20$) was significantly higher than for the traditional regression models ($M = 0.26$, $SD = 0.21$). The metamodels not only achieve higher correlation, but with smaller variation across folds. Therefore, they are both more effective and more stable than the traditional models.

Though the direct prediction of ASD behavioral score has rarely been studied, post-hoc analysis of classification models may correlate ASD behavioral scores with the classification scores to justify the model's efficacy. For example, a Pearson correlation coefficient of 0.348 between the classification scores and ADOS total score (the summation of ADOS communication and social scores) was reported in post-hoc analysis of a classification model [16]. Though not directly comparable, note that for the ADOS communication score we achieved up to 0.398 correlation using *predictive* analysis.

The generalization experiments show that the metamodel has better or similar generalizability compared to traditional regression models (Table 4). For the first generalization

**Table 3**. Pearson correlation coefficient between true and predicted scores from 8-fold cross-validation with Dataset A

| Base algorithm | $R_a$ mean(SD) | $M_a$ mean(SD) |
|---|---|---|
| LSTM | 0.2455(0.24) | 0.2693(0.20) |
| MLP  | 0.2723(0.22) | 0.3981(0.19) |
| SVM  | 0.3037(0.20) | 0.3582(0.19) |
| RF   | 0.2192(0.23) | 0.3077(0.21) |
| LR   | 0.2528(0.23) | 0.2915(0.21) |

**Table 4**. Pearson correlation coefficient between true and predicted scores for Dataset B

| Base algorithm | $R_a$ on dataset B mean(SD) | $M_a$ on dataset B mean(SD) | $R_b$ on $TS_b$ mean(SD) | $M_b$ on $TS_b$ mean(SD) |
|---|---|---|---|---|
| LSTM | 0.1693(0.05) | 0.2322(0.06)* | 0.1418(0.25) | 0.2599(0.12) |
| MLP  | 0.0326(0.04) | 0.1105(0.05)* | 0.0685(0.23) | 0.0459(0.16) |

* Significantly different compared to $R_a$, paired two-tailed t-test with $p < 0.05$

method where the previously trained models are directly applied, the metamodels $M_a$ performed better than the traditional regression models ($R_a$) on average for both LSTM and MLP (paired two-tailed t-test, $p = 0.03$ and $p < 0.001$, respectively). Furthermore, the LSTM $M_a$ result produced a significant correlation value ($r(110) = 0.2322$, $p = 0.01$). However, the performance of the newly trained meta-level model $M_b$ varies across different base algorithms: LSTM-based $M_b$ outperforms $R_a$ while MLP-based $M_b$ performs similarly to $R_a$. Thus, the success of the metamodel structure is dependent on the underlying learning approach.

We also compared traditional regression models $R_b$ trained on $TR_b$ and optimized on $VS_b$. The LSTM-based $M_b$ also achieved higher correlation than $R_b$ on average. In addition, a significant correlation between all predictions and the true scores for dataset B is observed for LSTM-based $M_b$ ($r(110) = 0.2308$, $p = 0.02$), while the correlation for the whole dataset for $R_b$ is not significant ($r(110) = 0.1068$, $p = 0.27$). Thus, using the pretrained base models and training just the meta-level model on a new dataset is capable of learning more predictive models than training traditional regression models on the new data from scratch.

## 4. CONCLUSIONS

We proposed a metamodel structure for regression problems where classification base models are first used to learn varying information, followed by a meta-level model which combines the base model information and produces the prediction. The metamodel was tested on the prediction of ADOS_COMM from rsfMRI using 5 base algorithms. The metamodel showed promise in increasing the correlation between true and predicted scores compared to traditional regression methods, on both cross-validation and new data generalization experiments. To further improve the prediction result, we could test this metamodel structure on more classification methods and potentially combine multiple learning algorithms to better capture different features for generating the prediction. Different algorithms could also be explored for the meta-level model.

## 5. COMPLIANCE WITH ETHICAL STANDARDS

This study was conducted retrospectively using human subject data made available in open access by the Autism Brain Imaging Data Exchange at http://fcon_1000.projects.nitrc.org/indi/abide/. Ethical approval was not required as confirmed by the license (CC BY-NC-SA 3.0) attached with the open access data.

## 6. ACKNOWLEDGMENTS

The authors have no relevant financial or non-financial interests to disclose.